\documentclass[10pt,twocolumn,letterpaper]{article}

\usepackage[pagenumbers]{cvpr} 

\definecolor{cvprblue}{rgb}{0.21,0.49,0.74}
\usepackage[pagebackref,breaklinks,colorlinks,allcolors=cvprblue]{hyperref}

\usepackage{marvosym}
\usepackage{multirow} 
\usepackage{bibunits}
\usepackage{tikz}
\usepackage{shadow}
\usepackage{rotating} 
\usetikzlibrary{matrix}
\usetikzlibrary{shadows.blur, shadows, positioning}
\usepackage{pifont}
\newcommand{\cmark}{\ding{51}}%
\newcommand{\xmark}{\ding{55}}%

\def\paperID{*****} 
\def\confName{CVPR}
\def\confYear{2025}

\title{Can Visual Encoder Learn to See Arrows?}

\author{
Naoyuki Terashita$^{1}$\thanks{E-mail: \href{mailto:naoyuki.terashita.sk@hitachi.com}{\texttt{naoyuki.terashita.sk@hitachi.com}}}\quad
Yusuke Tozaki$^{1,2}$\thanks{This work was done while the authors were interns at Hitachi, Ltd.}\quad
Hideaki Omote$^{1,3}$\footnotemark[2]\quad
Congkha Nguyen$^{1}$\\
Ryosuke Nakamoto$^{1}$\quad
Yuta Koreeda$^{1}$\quad
Hiroaki Ozaki$^{1}$\\
$^{1}$Hitachi, Ltd.\quad
$^{2}$Kyoto Sangyo University\quad
$^{3}$Gifu University
}

\begin{document}
\maketitle
 \begingroup
  \renewcommand\thefootnote{}
  \footnotetext{This work has been accepted for poster presentation at the \textit{Second Workshop on Visual Concepts in CVPR 2025}.}%
  \addtocounter{footnote}{-1}
\endgroup
\begin{abstract}
The diagram is a visual representation of a relationship illustrated with edges (lines or arrows), which is widely used in industrial and scientific communication.
Although recognizing diagrams is essential for vision language models (VLMs) to comprehend domain-specific knowledge, recent studies reveal that many VLMs fail to identify edges in images.
We hypothesize that these failures stem from an over-reliance on textual and positional biases, preventing VLMs from learning explicit edge features.
Based on this idea, we empirically investigate whether the image encoder in VLMs can learn edge representation through training on a diagram dataset in which edges are biased neither by textual nor positional information.
To this end, we conduct contrastive learning on an artificially generated diagram--caption dataset to train an image encoder and evaluate its diagram-related features on three tasks: probing, image retrieval, and captioning.
Our results show that the finetuned model outperforms pretrained CLIP in all tasks and surpasses zero-shot GPT-4o and LLaVA-Mistral in the captioning task.
These findings confirm that eliminating textual and positional biases fosters accurate edge recognition in VLMs, offering a promising path for advancing diagram understanding.
\end{abstract}

\begin{figure*}
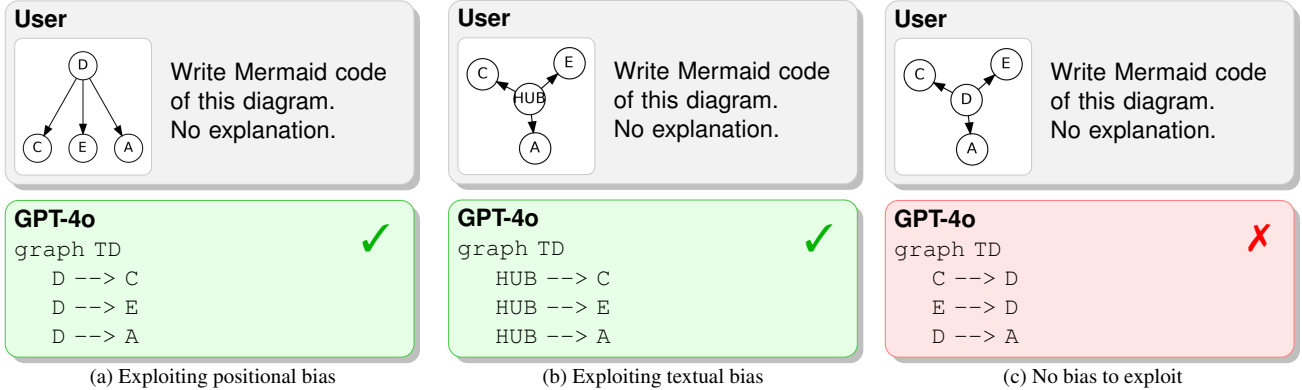

    \begin{subfigure}{0.33\linewidth}   
        \centering
        \input{figs/main_c.tex}
        \caption{Exploiting positional bias}
        \label{fig:sample_tree}
    \end{subfigure}
    \hfill
    \centering
    \begin{subfigure}{0.33\linewidth}
        \centering
        \input{figs/main_b.tex}
        \caption{Exploiting textual bias}
        \label{fig:sample_hub}
    \end{subfigure}
    \hfill
    \begin{subfigure}{0.33\linewidth}
        \centering
        \input{figs/main_a.tex}
        \caption{No bias to exploit}
        \label{fig:sample_b}
    \end{subfigure}
    \caption{Examples of diagram captioning by GPT-4o \cite{gpt4o}: (a) inferring relationships based on conventional top-down hierarchies, (b) leveraging semantic relationships between node labels, and (c) struggling when neither positional nor textual biases are available. All results were produced by \texttt{gpt-4o-2024-08-06} with temperature 0.}
    \label{fig:combined}
\end{figure*}

\section{Introduction}
Diagram is a simplified and structured visual representation of relationships using shapes connected by edges (lines or arrows).
Flowcharts, electronic circuits, and chemical structure diagrams are all examples of diagrams, and they play a major role in industrial and scientific communication.  
For a vision language model (VLM) to fully understand the context and knowledge in such domains, it is critical to accurately recognize diagram images.

However, recent studies suggest that VLMs might not accurately recognize edges in diagram images.
Yoshida et al. \cite{yoshida-etal-2024-how} have indicated that the feature representations of the CLIP \cite{radford2021learning} encoder, widely used in VLMs, may not contain sufficient information to classify the presence and direction of arrows in diagram images.
From a similar motivation, Rahmanzadehgervi et al. \cite{rahmanzadehgervi2024vision} proposed a VLM benchmark that requires the model to answer questions about lines and shapes, demonstrating even large VLMs such as GPT-4o \cite{gpt4o}  and Gemini-1.5 Pro \cite{team2024gemini} can sometimes fail on even simple questions. 

\begin{figure*}[t]
    \centering
    \begin{subfigure}{0.42\linewidth}
        \centering
        \includegraphics[width=\linewidth]{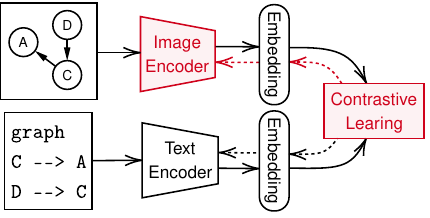}
        \caption{Training}
        \label{sub:train}
    \end{subfigure}
    \hfill
    \begin{subfigure}{0.53\linewidth}
        \centering
        \includegraphics[width=\linewidth]{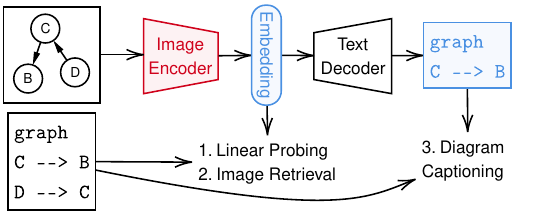}
        \caption{Evaluation}
        \label{sub:eval}
    \end{subfigure}
    \caption{Overview of our approach: (a) training a CLIP model with diagram--caption pairs that eliminate positional and textual biases, and (b) evaluating the model on three tasks: linear probing, image retrieval, and diagram captioning.}
    \label{fig:overall}
\end{figure*}

One reason VLMs often fail to recognize edges is that their visual training relies too heavily on positional or textual biases, hindering VLMs from learning edge features.
This can be demonstrated through a simple experiment shown in \cref{fig:overall}; GPT-4o succeeds in describing a diagram when it can rely on common-sense biases derived from node positions \subref{fig:sample_tree} or textual cues \subref{fig:sample_hub}, but fails when no such clues are available \subref{fig:sample_b}.
Recent benchmarks on visual math problems \cite{zhang2024mathverse} and flowchart VQA \cite{flowvqa2023} have also shown that VLMs tend to rely on textual and positional biases.

Based on these observations, this study experimentally demonstrates that eliminating textual and positional biases during training enables visual models to learn edge features.
To this end, we artificially generate a dataset of diagram images and text captions designed so that the presence and direction of edges cannot be inferred from text or position.  
We train CLIP, a common image encoder in VLMs, through contrastive learning on this dataset, then evaluate how well it captures edge information using three tasks: linear probing, image retrieval, and a newly proposed task called diagram captioning.
In the linear probing, we classify edge existence and direction with acquired features, while the image retrieval evaluates the model's ability to find images representing the identical graph with possibly different visual layout.
In our diagram captioning, we train a text decoder on the image encoder to predict the edge sets that appear in given diagram images.

Results from all three tasks show that our finetuned model substantially outperforms the original pretrained CLIP, indicating that our approach encourages acquiring edge representations invariant with textual and positional information.  
On our diagram captioning, the finetuned models also exceed the zero-shot performance of GPT-4o and LLaVA-Mistral, highlighting the current limitation of large VLMs and the effectiveness of our approach.

\section{Related Work}
Recently, large vision language models (LVLMs) have achieved human-level performance on a variety of VQA tasks \cite{gpt4o,team2024gemini,Chen2023PaLI-X,liu2023visual}, yet it has become clear that they rely heavily on textual content and positional layout to answer.

Chen et al.\cite{chen2024are} demonstrated that GeminiPro \cite{team2024gemini} solves 42.90\% of MMMU tasks \cite{yue2024mmmu} without any image inputs.
This reveals that existing results of common VQA benchmarks might not reflect the actual vision capability of LVLMs.
Further, in visual math benchmarks \cite{chen2021geoqa,lu2024mathvista,yue2024mmmu}, removing problem texts regarding visual information substantially drops performance \cite{zhang2024mathverse}, indicating that many LVLMs merely rely on textual information.
The limited capability in figure recognition is further illustrated by Rahmanzadehgervi et al.\ \cite{rahmanzadehgervi2024vision}, who reveal that even state-of-the-art models like GPT-4V \cite{achiam2023gpt} fail at simple visual tasks such as counting overlapping shapes or determining line segment intersections.
LVLMs also tend to rely on layout information.
In a flowchart VQA task, simply flipping the layout vertically significantly degrades performance \cite{flowvqa2023}.

In this work, we show that by removing biases tied to text or positional information, VLMs can learn to recognize lines and arrows purely from visual inputs.

\section{Learning with Debiased Diagram Dataset}
As shown in \cref{fig:overall}, our approach consists of artificially generating diagram--text pairs that exclude text and positional biases (\cref{sec:dataset}), followed by contrastive learning to finetune CLIP (\cref{sec:train}).  

\subsection{Diagram Dataset without Positional and Textual Biases}
\label{sec:dataset}
We aim to build an image dataset with captions that eliminate biases arising from text and positional information.
To have such a dataset with sufficient diversity, we generate diagram images and their Mermaid-style captions from randomly generated directed graphs.
Note that we use "graph" to refer to the abstract mathematical structure and "diagram" to denote its visual representation.

Each sample in our dataset pairs an image with text representing a directed graph containing different numbers of alphabet-labeled nodes.  
The directed graphs are generated so that their edges are generated independently with a fixed probability for each pair among eight nodes, excluding self-loops and bidirectional edges. 
When a generated graph has more than one weakly connected component, we keep only the largest one, having graphs with different numbers of nodes (two to eight).
For each graph, we draw a diagram image as in \cref{fig:overall} whose node positions are laid out by the force-directed placement \cite{fdp} with a random initial node layout.
This initialization ensures that the same graph can produce diagram images with different layouts. 
The captions describe the generated directed graph in Mermaid format, where each line denotes a directed edge; e.g., \texttt{A~-->~B} indicates an edge from node ``A'' to node ``B''.  
We generate 100k image--caption pairs and use 10\% of them as a test set.

\subsection{Training Encoders via Contrastive Learning}
\label{sec:train}
We finetune pretrained CLIP models using contrastive learning on our artificially generated dataset. 
CLIP \cite{radford2021learning} is a dual-encoder architecture, comprising an image and text encoder, that learns joint representations from pairs of images and their captions.
During training, CLIP minimizes a contrastive loss that brings the embeddings of matching image--text pairs closer while pushing apart the embeddings of non-matching pairs.  
We specifically target CLIP for our approach because its image encoders serve as foundational components in numerous state-of-the-art large vision language models \cite{zhu2023minigpt4,li2023blip2,liu2023visual,awadalla2023openflamingo}.

For our experiments, we adopt two pretrained CLIP models with different image encoder sizes: CLIP-ViT-B/32 \cite{clip-vit-base-patch32} and CLIP-ViT-L/14 \cite{clip-vit-large-patch14-336}.
CLIP-ViT-B/32 has 12 hidden layers and outputs 512-dimensional embeddings, whereas CLIP-ViT-L/14 has twice as many hidden layers and outputs 768-dimensional embeddings.
We implement standard contrastive learning using our artificial dataset with a sufficiently large number of training steps.

\begin{table*}[t]
\centering
\caption{Performance comparison of pretrained and finetuned CLIP models on diagram understanding tasks.}
\label{tab:main}
\begin{tabular}{l c c c c c}
\toprule
\multirow{2}{*}{Method} & \multicolumn{3}{c}{Linear Probing (Mean Accuracy)} & \multicolumn{2}{c}{Image Retrieval} \\
\cmidrule(lr){2-4} \cmidrule(lr){5-6}
                        & Node existence   & Edge existence & Edge direction & MAP@100 & MRR@100 \\
\midrule
Random    & 0.500 & 0.500 & 0.500 & 0.0004 & 0.001\\
\midrule
Pretrained ViT-B/32   & 0.959 & 0.639 & 0.518 & 0.067  & 0.108 \\
Pretrained ViT-L/14   & 0.999 & 0.725 & 0.509 & 0.131  & 0.170 \\
\textbf{Finetuned ViT-B/32}   & 0.994 & 0.726 & 0.857 & 0.973  & 0.973 \\
\textbf{Finetuned ViT-L/14}   & \textbf{1.000} & \textbf{0.735} & \textbf{0.860} & \textbf{0.996}  & \textbf{0.996} \\
\bottomrule
\end{tabular}
\end{table*}

\section{Evaluation of Image Encoder}
\label{sec:eval}
We evaluate the finetuned image encoder on three tasks that rely on diagram recognition: the linear probing (\cref{sec:probing}), image retrieval (\cref{sec:retrieval}), and diagram captioning (\cref{sec:GPT-2}).

\subsection{Linear Probing}
\label{sec:probing}
Linear probing \cite{alain2016understanding,conneau-etal-2018-cram} measures how well the extracted features encode information through classification tasks on features.
In this study, we train and evaluate a simple logistic regression on top of the frozen image encoder to quantify the recognition capability of nodes and edges.

We define three binary classification tasks: \emph{node existence}, \emph{edge existence}, and \emph{edge direction} classification.  
In node existence classification, for a given node label (``A'' to ``H''), we predict whether it appears in the diagram.  
Edge existence classification is a task to predict if an undirected edge exists between a given pair of nodes (ignoring direction).  
If either node is missing from the graph, that test sample is excluded to purely evaluate the edge recognition ability.
Finally, in edge direction classification, we check whether a specific directed edge exists (e.g., from A to B). 
If no edge exists in either direction between the given two nodes, we skip that sample to solely evaluate the direction-related performance.
We compute the accuracy for every possible label, node pair, or directed edge to report the average.  
For all tasks, we use balanced undersampling to ensure that the accuracies of all tasks can be compared to the chance rate (0.5).

To see the effect of our additional contrastive learning, we adopt pretrained CLIP-ViT-B/32 and CLIP-ViT-L/14 without additional contrastive training as baselines.
As shown in \cref{tab:main}, the pretrained baseline models perform poorly at edge direction classification (roughly at chance level), although they excel at text recognition (node labels).  
In contrast, both finetuned models show significant improvements from their baselines, especially in edge-direction classification (e.g., ViT-L/14 jumps from near-chance to 86\% accuracy). 
These results indicate that removing textual and positional biases via contrastive learning lets the image encoder acquire edge-related features.

\subsection{Image Retrieval}
\label{sec:retrieval}
Our image retrieval task requires the model to retrieve all diagram images that represent the same directed graph as a given query image, which falls within the broader task category called content-based image retrieval \cite{sharif2014cnn,babenko2014neural}.
This task tests whether the learned features are invariant to node positions, as the query and target diagrams can have different layouts while representing identical graph structures.

We newly generate 1,000 query graphs using the same method in \cref{sec:dataset} ensuring each query graph appears in our test dataset (but possibly with a different layout).  
We encode both the query and all test images with the same image encoder, rank them by cosine similarity, and measure Mean Average Precision (MAP) and Mean Reciprocal Rank (MRR) up to the top 100 results.

\begin{figure}[t]
    \centering
    \input{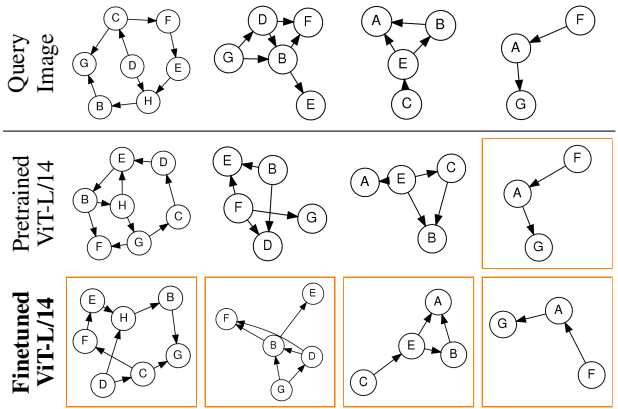}
    \caption{Examples of query images (top row) and the top retrieved images using the pretrained ViT-L/14 (middle row) and finetuned ViT-L/14 (bottom row). Images surrounded by orange lines represent true positives that share the same directed graph structures as the queries.}
    \label{fig:retrieved}
\end{figure}

As shown in \cref{tab:main}, our finetuned ViT-B/32 and ViT-L/14 achieve MAP and MRR scores above 0.97, indicating that they successfully learn diagram features that are invariant to node positions.
\cref{fig:retrieved} shows examples of query images and retrieval results from finetuned and pretrained ViT-L/14.
The pretrained image encoder mostly tends to focus on text and layout similarity, and it succeeds only when the node layouts are extremely similar.
By contrast, the finetuned models correctly retrieve matching graphs even if the node layouts differ significantly.

\subsection{Diagram Captioning}
\label{sec:GPT-2}
This section proposes a new task called diagram captioning that requires a VLM to describe the edges in a given diagram image, accompanied by training a text decoder.

Our diagram captioning is a task to predict a Mermaid-style description of the diagram presented in an input image.
Performance is measured by the micro F1-score of the predicted edge set, obtained by parsing edge descriptions in the predicted Mermaid text (e.g., \texttt{A --> B}).

To construct our VLM, we pair CLIP's image embeddings with GPT-2 \cite{radford2019language,hf_canonical_model_maintainers_2022} as a text decoder. 
Our GPT-2 uses cross-attention on the image embeddings and previous tokens, predicting the next token probabilities. 
We train on our artificial image--caption dataset with cross-entropy loss, freezing the image encoder's weights, and select checkpoints based on validation loss from 1\% of the training set. 
We compare the finetuned models with baselines that use the original CLIP encoders for image features, as well as with zero-shot inference from large VLMs, namely GPT-4o \cite{gpt4o} (\texttt{gpt-4o-2024-08-06}) and LLaVA-Mistral \cite{liu2024llavanext} (which also uses CLIP-ViT-L/14).
Zero-shot inference is prompted to generate a Mermaid-format caption describing the given diagram image.

\cref{tab:vlm} shows that the finetuned ViT-L/14 encoder achieves an F1 of 0.966, outperforming pretrained ViT-L/14 and clearly beating GPT-4o with zero-shot.  
This indicates that the improved edge representation confirmed by the linear probing and image retrieval also benefits practical downstream tasks.
The lower performance of the pretrained ViT-L/14 in both the diagram captioning and linear probing (\cref{tab:main}) indicates that simply adapting the decoder is not enough; the bottleneck lies in the image encoder.
GPT-4o and LLaVA-Mistral were shown to struggle with tasks that have no textual and positional cues to rely on, which is consistent with our findings in \cref{fig:overall}.
Although a supervised instruction tuning on these models would likely improve performance and provide insightful results, we leave this for our future work.
We also tested our models on diagram images whose graphs are non-isomorphic to any training sample (even as unlabeled and undirected). 
    Despite a slight performance degradation, our finetuned models still significantly outperformed the baselines, showing their strong generalization to unseen graph structures.
    
    \begin{table}[t]
    \centering
    \caption{Diagram captioning performance comparison.}
    \label{tab:vlm}
    \begin{tabular}{l c}
    \toprule
    Method & F1-score \\
    \midrule
    Llava-Mistral \cite{liu2024llavanext}            & 0.118 \\
    GPT-4o \cite{gpt4o}                   & 0.500 \\
    \midrule
    Pretrained ViT-B/32 (+GPT-2)            & 0.413 \\
    Pretrained ViT-L/14 (+GPT-2)            & 0.668 \\
    \textbf{Finetuned ViT-B/32 (+GPT-2)}    & 0.516 \\
    \textbf{Finetuned ViT-L/14 (+GPT-2)}    & \textbf{0.966} \\
    \bottomrule
    \end{tabular}
    \end{table}
    
\section{Conclusion}
We showed that removing textual and positional biases enables VLMs to learn edge recognition in diagrams. 
Using a synthetic dataset and contrastive learning on CLIP-based encoders, our finetuned models outperformed pretrained baselines across linear probing, image retrieval, and diagram captioning.
This highlights the effectiveness of removing textual and positional biases for teaching VLMs to capture diagram structure.

\small
\bibliographystyle{ieeenat_fullname}
\bibliography{main.bib}

\end{document}